\documentclass[runningheads]{llncs}

\usepackage{silence}
\usepackage[T1]{fontenc}
\usepackage{amsmath}
\usepackage{graphicx} 
\usepackage{multirow} 
\usepackage{amssymb,amsfonts}%
\usepackage{booktabs}%

\usepackage{adjustbox}
\usepackage[colorlinks=true, allcolors=blue]{hyperref}
\usepackage{enumitem}
\usepackage{subcaption}
\usepackage{caption}
\usepackage{bbding} 
\usepackage[utf8]{inputenc}

\begin{document}

\captionsetup[table]{skip=8pt}

\title{Feature Attribution-Based Explainability Analysis of Deep Learning Models in Predictive Process Monitoring}

\author{Kseniya Sahatova\inst{1, 2}\textsuperscript{\Envelope}\orcidID{0009-0001-0301-6361} \and Rafael Seidi Oyamada \inst{2}\orcidID{0000-0002-4408-9575} \and 
Xuefei Lu\inst{1}\orcidID{0000-0003-2103-6478} \and Johannes De Smedt\inst{2}\orcidID{0000-0003-0389-0275}}

\institute{SKEMA Business School, 92156 Suresnes, France \\ \email{kseniya.sahatova@skema.edu, xuefei.lu@skema.edu} \and
KU Leuven, B-3000 Leuven, Belgium \\ \email{rafael.oyamada@kuleuven.be, johannes.desmedt@kuleuven.be}}

\authorrunning{K. Sahatova et al.}
\titlerunning{Feature Attribution-Based Explainability Analysis in PPM}

\maketitle    % typeset the header of the contribution
\begin{abstract}
% Predictive process monitoring (PPM) supports the  optimization and control of operational business processes by enabling tasks such as predicting the outcome of an ongoing process instance, estimating its remaining time, or forecasting the next sequence of activities. Recurrent neural networks, namely long short-term memory (LSTM) models, have demonstrated strong performance in these tasks due to their ability to capture sequential and temporal dependencies inherent in event logs.
% However, despite their predictive power, LSTM-based models remain opaque, which limits their trustworthiness and practical adoption. To address this issue, feature attribution methods have been widely explored.
Predictive process monitoring supports the optimization and control of operational business processes by forecasting the future state or outcome of ongoing cases.
%such as predicting the outcome of an ongoing process instance, estimating its remaining time, or forecasting the next sequence of activities. 
While deep neural networks have achieved strong performance for these tasks by modeling sequential dependencies in event logs, their black-box nature limits trust and practical adoption. Feature attribution methods are often used to address this, but applying them directly poses a dilemma: event-level attributions impose high computational complexity for long traces, while explanations based on aggregated trace representations often fail to capture the underlying control-flow dynamics. %, motivating the use of feature attribution methods. Although their straightforward application\textemdash either at the individual event level or on aggregated trace representations\textemdash poses significant challenges, event-level feature attributions impose high computational complexity for long traces, while explanations for the aggregated case representations might not reproduce the relations in the underlying control flow.
To address this issue, we propose a local post-hoc explainability method for deep neural networks in outcome prediction. The method relies on a control-flow-aware segmentation algorithm that partitions a trace into meaningful segments and supports the computation of segment-level SHAP explanations. This makes it possible to identify which parts of a trace influence a prediction and which change points steer the case toward the predicted outcome. We assess the proposed segmentation method on a synthetic dataset with known process logic, where meaningful change points can be explicitly verified, and we demonstrate its usefulness on real-world event logs from a loan application process and an administrative process of a Dutch municipality.
% In this paper, we propose a method for generating local post-hoc explanations for deep  neural networks applied to outcome prediction tasks in PPM. We introduce a control-flow-aware segmentation algorithm that partitions a process trace into meaningful segments, enabling the computation of segment-level SHAP explanations. This approach not only identifies which case segments influence the model’s predictions, but also highlights change points that steer the control-flow toward the predicted outcome. We evaluate the quality of the proposed segmentation method using a synthetic dataset with known business logic, where meaningful change points can be explicitly identified. Furthermore, we demonstrate the applicability of our approach by generating local explanations on real-world event logs for the loan application process and administrative process in a Dutch municipality.  

% \todo[inline]{Blackbox DL models commonly used in PPM --> black box nature calls for transparency --> commonly interesting results are event-level or for aggregated traces --> both explanations suffer from 1) the dependence structure in events 2) computational burden. To address the challenges, we propose a pipeline to generate segment-wise explanation.  This approach not only identifies which case segments influence the model’s predictions, but also highlights change points that steer the control-flow toward the predicted outcome. --> the proposed method has been applied on two case studies on loan application processes. }
\keywords{Predictive Process Monitoring  \and SHAP \and Control-flow-aware Trace Segmentation.}
\end{abstract}

\section{Introduction}

Businesses increasingly use Artificial Intelligence (AI) to anticipate issues such as delays or failures, enabling optimization and prevention. Predictive process monitoring (PPM), a branch of Process Mining (PM), focuses on predicting the outcome of an ongoing business process instance, estimating the remaining time before completion, or forecasting the sequence of activities in the future. To address these predictive tasks, researchers have adopted a variety of sequence modeling architectures~\cite{weytjens2020process,tax2017predictive}, such as recurrent neural networks (RNNs) and transformers, which have emerged as a dominant paradigm due to their ability to capture the sequential and temporal dependencies inherent in event logs.

Despite their strong predictive performance, sequential models in PPM remain largely obscure: their internal representations are difficult to interpret, limiting practical adoption in high-stakes operational settings. This challenge has motivated growing interest in explainability methods that facilitate understanding of how models trained on event logs arrive at their predictions \cite{bento_timeshap_2021,buliga2025guiding,galanti2020explainable,jiang2024seqshap,stevens2025generating}. Recent studies have explored the use of feature attribution techniques such as LIME (local interpretable model-agnostic explanations) and SHAP (Shapley additive explanations) to enhance the transparency of deep learning models in PPM \cite{elkhawaga_2022_xai,galanti2020explainable,rizzi2020explainability,velmurugan2021evaluating}. However, current approaches still explain predictions either at the level of individual events or at the level of the full aggregated trace. Both views are problematic for PPM: event-level explanations are often too fragmented to reveal meaningful process behavior, whereas trace-level explanations operate at a broader level of abstraction and introduce information loss through aggregation, making it harder to identify which part of the execution drives the prediction. Consequently, existing methods may assign importance scores without preserving the local control-flow context that gives those scores process meaning.

Event logs exhibit sequential and temporal dependencies, which makes them partly related to time-series data. This connection is relevant because the time-series literature has proposed several feature attribution methods that improve interpretability by grouping adjacent observations, in contrast to interpreting shallow event-level or fully aggregated data, into segments \cite{crabbe2021explaining,doddaiah2022class,liu2024explaining,schlegel2021ts,sivill2022limesegment}. 
In coalition-based methods such as Shapley values, this strategy also reduces the computational cost, which otherwise grows exponentially with the number of timesteps. However, event logs are not generic time series: they contain process-specific information such as control-flow behavior, activity semantics, and case progression, and are irregular in time, since events are recorded upon activity execution rather than at fixed sampling intervals as in regular time series. As a result, segmentation strategies developed for time series may produce groups of events that do not correspond to meaningful process fragments, for example by splitting a coherent process stage or merging unrelated control-flow behavior.

To address this gap, we propose a control-flow-aware segmentation algorithm that improves the computational efficiency of feature attribution-based explanations while preserving the control-flow constraints of business processes. Our method partitions traces into coherent subsequences of activities by identifying shifts in transition probabilities, thereby yielding more interpretable units. We compute segment-level SHAP values for sequential classifiers in a binary prediction setting. For stakeholders and practitioners, event-level feature attributions provide limited interpretation because the impact of an activity on the outcome often depends on its execution context. As a result, attributing predictive importance to individual activities may provide limited insight in isolation. We argue that computing SHAP values at the segment level groups activities that share similar transition dynamics, and that the underlying transition probabilities within each segment offer additional interpretive context beyond the attribution scores themselves.

The paper is organized as follows. Section \ref{sec:background} reviews existing work on explainability in PPM and motivates our study. Section \ref{sec:preliminaries} introduces the main definitions required to facilitate understanding of the methods used in this paper. Section \ref{sec:methodology} describes the proposed control-flow-aware segmentation algorithm and the metrics used for evaluating segmentation quality. Sections \ref{sec:setup} and \ref{sec:results} introduce the experimental setup and present the interpretation of the obtained results, while Section \ref{sec:discussion} discusses the findings summarized in Section \ref{sec:conclusion}.

\section{Related Works}
\label{sec:background}

Various explainability methods have been applied to enhance transparency and improve understanding of predictive models in process monitoring. In the work \cite{rizzi2020explainability}, LIME explanations were used to improve outcome prediction. Event-level feature attributions for false positives and false negatives were analyzed with rule mining to identify regions contributing to errors. The study  \cite{galanti2020explainable} shows how Shapley values can be adapted to explain an LSTM model in the remaining-time prediction task generating explanations for each feature per timestep. In the study \cite{velmurugan2021evaluating} the stability of LIME and the model-specific SHAP value estimation methods are examined in the outcome classification task across different types of trace encoding. Similar research \cite{elkhawaga_2022_xai} estimates the effect of design choices in predictive models and pre-processing techniques on global explainability methods, examining SHAP values through the dependence plot representation. In the study \cite{bento_timeshap_2021}, KernelSHAP is adapted to explain RNNs at the feature, event, and cell levels, where a cell corresponds to a specific feature of a specific event; that is, explanations are generated for each feature within a subset of events. To enable efficient computation for long sequences, the authors introduce a pruning algorithm based on the assumption that only recent events are relevant. Older events are therefore grouped into a single coalition and treated as less important for fine-grained explanations.

The previously described approaches compute explanations for a single event and its associated features, proposing algorithms for efficient computation while disregarding the correlations between adjacent events inherent in event logs. These temporal dependencies make event logs analogous to time series, enabling the adaptation of explainability approaches originally developed for time-series data to the field of PM. 
Active research on the explainability of sequential models for time series has introduced various techniques that identify semantically meaningful segments and subsequently generate explanations for them. The LIMESegment framework \cite{sivill2022limesegment} was developed to explain time-series segments based on shape patterns and statistical properties. TS-MULE \cite{schlegel2021ts} utilizes LIME to explain segments of univariate and multivariate time series, employing matrix profile and SAX transformations for segmentation.  Unlike in time series, consecutive activities in process traces may exhibit strong correlations that indicate structured process behavior, or they may show little to no correlation, suggesting loops, branching, or unpredictable behavior—i.e., process unstructuredness. 

WindowSHAP method \cite{nayebi2023window} was introduced to reduce computational complexity and variance in point-wise explanations. In this approach, adjacent data points are grouped using windowing techniques with fixed- and variable-length time windows. Jiang et al. \cite{jiang2024seqshap} adapt KernelSHAP for subsequence-level explanations by relying on the definition of sessions in recommendation system, where events can be grouped based on the time windows in which they occur. While such an approach is suitable for data, which temporal patterns are clearly distinguishable, it has limitations in identifying meaningful subsequences when the primary analytical focus is on the process control flow\textemdash i.e., when subsequences need to be grouped based on the types of activities performed.
While segment-based explanation methods have shown promise in the time-series domain, their direct application to event logs disregards the process-specific dependencies. This motivates our work, which bridges control-flow-aware trace segmentation with feature attribution to produce explanations aligned with the control-flow constraints of business processes

% Discussions within the xAI community concerning the representativeness of Shapley values (\cite{kumar2020problems}, \cite{chen2023algorithms}, \cite{sundararajan2020many}, \cite{janzing2020feature}) as a feature-attribution method have highlighted several key challenges. First, the application of Shapley values to ML models requires a well-defined notion of a coalition, as the size of feature subsets can vary and such models cannot accept inputs of arbitrary size. Thus, the subset of ``missing" features $\bar{S}$ needs to be replaced by some non-informative values.   
% \todo[inline]{Violation of feature independence assumption}
% \textit{Imputing missing features.} As mentioned, the imputation of dropped features that are not supposed to participate in a ``game" (execution of a model for the prediction obtaining) is a challenging task. The two main approaches established in the literature are \textit{marginal distribution} (referred to as interventional method) or \textit{conditional distribution} (referred to as conditional method) for sampling of replacement values of missing features. Kumar et al. \cite{kumar2020problems} raise the following challenge for a choice between these two: the interventional method tends to produce ``off-manifold" samples, whereas the conditional method assigns importance score to the feature that might have no influence to the prediction.  

\section{Preliminaries}
\label{sec:preliminaries}

In this section, we introduce the main concepts from predictive process monitoring and feature attribution used throughout this work.

\begin{definition}[Event, Trace, Prefix, Event Log]
Let $\mathcal{A}$ be the universe of activities and $\mathcal{C}$ the universe of case identifiers. An \textit{event} is defined as a tuple $e = (a, c, t, D)$, where $a \in \mathcal{A}$ is the activity, $c \in \mathcal{C}$ is the case identifier, $t$ is the timestamp, and $D$ is a set of event- or case-level attributes. We denote the universe of all possible events by $\mathcal{E}$.

A \textit{trace} $\sigma = \langle e_1, e_2, \dots, e_n \rangle$ is a non-empty ordered sequence of events such that $e_i \in \mathcal{E}$ for all $1 \leq i \leq n$, and $t_i \leq t_{i+1}$ for all $1 \leq i < n$. Given a trace $\sigma = \langle e_1, \dots, e_n \rangle$ and an integer $1 \leq m \leq n$, the prefix of length $m$ is defined as $\mathit{prefix}(\sigma,m) = \langle e_1, \dots, e_m \rangle$. The set of all possible traces over $\mathcal{E}$ is denoted by $\mathcal{E}^*$. Finally, an \textit{event log} $L$ is a multiset of traces.
\end{definition}

% In predictive process monitoring, predictions for an ongoing case are made using only its first $m$ events. This partial trace is called a \textit{prefix}. 

% \begin{definition}[Prefix] 
% Given a trace $\sigma = \langle e_1, \dots, e_n \rangle$ and an integer $1 \leq m \leq n$, the prefix of length $m$ is defined as $\mathit{prefix}(\sigma, m) = \langle e_1, \dots, e_m \rangle$.
% \end{definition}

In this work, we consider binary outcome prediction only, that is, the task of predicting the final outcome of an ongoing case from a trace prefix.

\begin{definition}[Outcome Classifier]
A binary outcome classifier is a function $f_{cl}: \mathcal{E}^* \to \{0,1\}$. In practice, we consider its associated score function $f_{cls}: \mathcal{E}^* \to [0,1]$, which maps a prefix $\sigma_{\leq m} = \mathit{prefix}(\sigma,m)$ to the estimated probability of a positive outcome.
\end{definition}

Our goal is to explain how a classifier arrives at its final prediction. To this end, we compute SHAP values over the segments produced by our proposed directly-follows-relation-based segmentation algorithm. 

\begin{definition}[Directly-Follows Relation (DFR)]
Let $\sigma = \langle e_1, \dots, e_n \rangle$ be a trace and let $x,y \in \mathcal{A}$ be two activities. We say that $x$ is in a directly-follows relation with $y$, denoted by $(x,y)$, if there exist two consecutive events $e_i$ and $e_{i+1}$ in $\sigma$ such that the activity of $e_i$ is $x$ and the activity of $e_{i+1}$ is $y$.
\end{definition}

\begin{definition}[{Shapley Values}]  Let $N=\{1,...,n\}$ be a set of features. Let $v: 2^N \to \ \mathbb{R}$ be a value function that assigns to each subset of features $S \subseteq N$ a real number $v(S)$, representing the model output when only features in $S$ are used. The Shapley value $\phi$ of the feature $i$ quantifies the weighted average marginal contribution to the model prediction over all possible feature subsets as:
\[
\phi_{i}(v) = \sum_{S \subseteq N\setminus \{i\}}\frac{|S|! (|N|-|S|-1)!}{|N|!} (v(S \cup \{i\}) - v(S))
\]
\end{definition} 

We utilize the following characteristic function: $v(S) = f_{cls}(x_S, x'_{N \setminus S}),$ where $x'$ is a baseline drawn from the training distribution and $f_{cls}$ is the scoring function. In outcome prediction, $v(N)$ corresponds to the classifier output $f_{cls}(\sigma_{\leq m})$. Shapley values satisfy several desirable properties, namely local accuracy,
missingness, and consistency; we refer to~\cite{aas2021explaining,lundberg_unified_2017,sundararajan2020many}
for further details.

\section{Segment-based Feature Attribution for Explainability}
\label{sec:methodology}
% In this section we introduce the inner workings of the segment-based feature attribution methodology.

We capture the evolution of process executions recorded in an event log through \emph{process phases}, in which consecutive events belong to the same local control-flow context. For example, in a loan process, a trace may contain a segment $\hat{\sigma}=\langle e_i, \dots, e_j \rangle$ ($1 \leq i < j \leq n$, where $n$ is the trace length) characterized as a segment of activities related to application completion, followed by a segment related to offer creation. The former might contain common transitions with high directly-follows probabilities, such as \texttt{Submit application} $\to$ \texttt{Check documents} and \texttt{Check documents} $\to$ \texttt{Validate application}. Later, when the case moves to the offer creation part, we may see a less common transition like \texttt{Validate application} $\to$ \texttt{Create offer}, characterizing a drop of likelihood that can mark the boundary between the two phases. Thus, a natural way to estimate such probabilities is through the discovery of a \emph{directly-follows graph} (DFG). 

More specifically, by obtaining the matrix of DFRs, we obtain empirical transition probabilities between activities. Evaluating these probabilities along the events of a trace yields a sequential transition segment that reflects the local control-flow consistency of the observed process execution. Positions where this segment changes can then be interpreted as boundaries between different process phases; we refer to such positions as \textit{change points}. Change point detection is a well-established task in time series analysis \cite{deldari2020espresso,hamed2026changepoint,li2024automatic,truong2020selective}, where the goal is to identify positions at which the statistical properties of a sequential signal change. In our case, the signal is given by the transition-probability profile derived from the event log, and a change point corresponds to a shift in the control-flow dynamics of the process instance, which can be interpreted as a phase boundary. Thus, let us introduce our concept of trace segmentation.

\begin{definition}[Trace Segmentation]
A \textit{trace segmentation algorithm} is a function $f_{seg}: \mathcal{E}^* \rightarrow \mathcal{S}^*$ that partitions a trace $\sigma$ into a sequence of segments $s = \langle s_1, s_2, \dots, s_k \rangle$.
Each segment $s_l \in \mathcal{S}$ is a sub-trace $s_l = \langle e_{i}, \dots, e_{m} \rangle$ where $1 \leq i \leq m \leq n$, such that the concatenation of all segments $s_1 \cdot s_2 \cdot \dots \cdot s_k$ equals the original trace $\sigma$.
\end{definition}

Using the introduced definition of trace segments and change points, we propose a control-flow-aware segmentation algorithm that identifies the corresponding change points and partitions a trace into segments. The algorithm optimizes a DFR-based cost function using dynamic programming, which we discuss further. Given these segments and a sequence modeling classifier, we provide explanations.

\subsection{Control-flow-aware Segmentation Algorithm}

Our control-flow aware segmentation algorithm combines (i) a cost function derived from directly-follows relations and (ii) the PELT (Pruned Exact Linear Time) algorithm~\cite{killick2012optimal}, which detects multiple change points by minimizing a penalized objective. 

First, from the event log, we discover a DFG, extract the DFR matrix, and normalize it to obtain the transition probabilities between activities. Subsequently, given a trace $\sigma$, the goal is to identify $k$ change points that partition the trace into $k+1$ segments. To assess whether a candidate segment is coherent, we assign it a cost based on the directly-follows probabilities. More specifically, transitions that are common in the event log contribute little to the cost, whereas rare or unseen transitions increase it. Therefore, segments that follow the typical control-flow behavior receive low cost, while segments containing unlikely behaviors receive high cost. Formally, for a segment $s=(a_i, \dots, a_j)$, we define: 

\begin{equation}
C(s) = \sum_{i=1}^{m-1} c(a_i, a_{i+1}),
\end{equation}
where
\begin{equation}
c(a_i, a_{i+1}) =
\begin{cases}
-\log T[a_i, a_{i+1}] & \text{if } T[a_i, a_{i+1}] > 0, \\
\gamma & \text{otherwise,}
\end{cases}
\end{equation}
where $T[a_i, a_{i+1}]$ denotes the directly-follows probability of activity $a_{i+1}$ after $a_i$, and $\gamma$ is some penalty for the transitions that have not been captured by the DFR matrix.

For example, if the transitions \texttt{Submit application} $\to$ \texttt{Check documents} and \texttt{Check documents} $\to$ \texttt{Validate application} are frequent in the event log, then the segment \texttt{Submit application} $\to$ \texttt{Check documents} $\to$ \texttt{Validate application} will receive a low cost. In contrast, a segment containing a rare or unseen transition will receive a higher cost, indicating a possible shift in local control-flow behavior. Therefore, PELT identifies the change points by minimizing the total segmentation cost over the trace. This objective combines the cost of the resulting segments with a penalty term for introducing additional change points:
\begin{equation}
\min_{\tau_{1:k}} \sum_{i=1}^{k+1} C\!\left(a_{(\tau_{i-1}+1):\tau_i}\right) + \beta k,
\end{equation}
where $C(\cdot)$ denotes the segment cost, $\tau_{1:k}$ are the detected change points, and $\beta$ controls the trade-off between segmentation quality and model complexity. This hyperparameter is selected via grid search by minimizing the Akaike Information Criterion (AIC), which balances goodness of fit against the number of change points, thereby favoring the less costly segmentation for a given trace. In particular, the penalty term discourages over-segmentation, such that new boundaries are introduced only when they lead to a substantial improvement in local control-flow consistency.

\vspace{-6pt}

\subsection{Calculation of Segment-level SHAP Values}

Trace segmentation allows us to explain predictions at the level of coherent trace fragments. This is more informative in process mining than explaining either single events or the entire trace at once. Event-level explanations often ignore the broader local control-flow context, whereas trace-level explanations collapse the whole process instance into one explanatory unit. By using segments instead, we can quantify how different parts of a trace, each corresponding to a locally consistent control-flow region, contribute to the prediction. In other words, each segment is treated as one interpretable process unit in the SHAP computation.
The calculation of segment-level SHAP values can be formalized as follows. 

Let $\sigma = \langle e_1, \dots, e_n\rangle $ be a trace of length $n$. We represent $\sigma$ as a matrix $X$ of the size $n \times d$, where the $i$-th row $x_i \in \mathbb{R}^d$ is the feature vector of event $e_i$, and $d$ is the number of features. Let $B$ be a baseline matrix of the same dimensions $n \times d$. Following the approach in \cite{bento_timeshap_2021}, we create a reference event vector $\bar{e} \in \mathbb{R}^d$ that summarizes typical behavior in the training data. For a categorical attribute $j$, $\bar{e}_j = \text{mode}(X_{*,j})$, where $X_{*,j}$ denotes the $j$-th column of $X$. For a numerical attribute $j$, $\bar{e}_j = \bar{X}_{*, j}$, representing the mean of the $j$-th column. Further, the baseline B is constructed as $B = \mathbf{1}_n\cdot \bar{e}$.
Suppose the trace is partitioned into $k$ contiguous segments $s_l, \ l = 1,\dots, k$.
To treat segments as explanatory units in the SHAP framework, we define a coalition vector $z \in \{0, 1\}^k$, where $z_l = 1$ indicates that segment $s_l$ retains its original values from $X$, and $z_l = 0$ indicates that it is replaced by the corresponding baseline values from $B$ .
For a given coalition $z$, we construct the mask matrix $M(z) \in \{0, 1\}^{n \times d}$, where $M(z)_{i, j} = z_l,  \forall i \in s_l, j=1,\dots,d$. The perturbation function $h$  is given as $h_s(z) = M(z) \odot X + (J-M(z)) \odot B$,
where $J  \in \mathbb{R}^{n \times d}$ is an all-ones matrix. Applying KernelSHAP with the $k$ segments as features, we obtain segment-level SHAP values $\phi_1, \dots, \phi_k$ satisfying the efficiency property $\sum^k_{l=1}{\phi_l} = f(X) - f(B)$, where $f(B)$ serves as the base value.

\section{Experimental Setup}
\label{sec:setup}

This section introduces the employed event logs, the predictive models for predictions, the segmentation baselines, and the evaluation protocol used to study segment-level explanations. The code for reproducing the results is available on GitHub \footnote{\url{https://github.com/kSahatova/segment-level-expl-PPM.git}}.

\subsection{Event Logs and Predictive Model}

We use one synthetic and two real-world benchmark logs. The synthetic log takes into consideration the challenge behind the evaluation of segmentation strategies due to the absence of ground-truth change points. Thus, we employ the SimBank simulator \cite{demoor2025simbank}, a tool for benchmarking prescriptive process monitoring methods. The simulator generates synthetic cases of a fictional loan application process and allows control over outcomes via a predefined process model and interventions. We select two samples with clearly defined change points from the simulator output\footnote{\url{https://zenodo.org/records/15574272}}. The first sample is built around the activity \textit{skip\_contact\_hq}, which leads to the \textit{cancellation} outcome. The second sample uses \textit{contact\_hq} and \textit{calculate\_offer}, which lead to the \textit{acceptance} outcome. These known positions are used as ground truth for evaluating segmentation quality.

Regarding the real-world logs, we employ the BPIC15\footnote{\url{https://data.4tu.nl/collections/BPI_Challenge_2015/5065424/1}} and the BPIC17\footnote{\url{https://data.4tu.nl/articles/dataset/BPI_Challenge_2017/12696884}}. For both, we adopt the outcome labeling strategies from the literature~\cite{teinemaa2019outcome}.
The former contains administrative processes from Dutch municipalities. We use the first municipality, and cases are labeled positive if they satisfy the LTL rule that whenever \textit{send confirmation receipt} occurs, it is eventually followed by \textit{retrieve missing data}; otherwise, it is labeled negative. 
On the other hand, the BPIC17 describes a loan application process from a financial institution, and we filter only the accepted and canceled cases. For both logs, in order to focus on predictions made before the outcome is explicitly observed, we extract fixed-length prefixes and remove traces that already contain outcome-revealing activities.

As a sequence modeling classifier, we use LSTM models because it is often used in the literature \cite{tax2017predictive}. Since our focus is not to optimize predictive performance but provide explanations from predictions, we opt only for this model and a simple hyperparameter optimization by sweeping the following values: the number of layers ${1, 2, 3}$, the embedding sizes ${32, 64, 128}$, the hidden sizes ${64, 128}$, and the learning rates ${1e-4, 5e-5, 1e-5}$. As data preprocessing, we exclude outlier cases falling outside the 5th to 95th percentile of all set cases, and use the activity labels and timestamp-related features for training. 
To evaluate model predictions before the outcome is determined, we extract prefixes of 30 events and exclude cases in which outcome-revealing activities have already occurred. Since the goal of PPM is to provide accurate predictions as early as possible during case execution, the selected prefix length balances predictive performance and explanatory value. Shorter prefixes often lack sufficient context for accurate predictions, whereas longer prefixes are more likely to contain outcome-revealing events, causing explanations to be dominated by these events and thus reducing their informativeness. The model trained on BPIC15 obtained 92\% AUC, and the model trained on BPIC17 achieved 85.14\%.

\subsection{Segment-level explanations and quantitative evaluation} 

\paragraph{Baseline explainability methods.} We compare our control-flow-aware segmentation against two baselines: the ``per-event'' explanation, which is the most common approach from the literature  \cite{galanti2020explainable}, and the ``distribution-based segmentation"~\cite{jiang2024seqshap}. The latter also proposed for identification of the most probable segments relying on the sessions definition in recommendation system. It depends on several hyperparameters: we set the minimum and maximum window sizes to 3 and 5, respectively, for the initial segmentation stage, and utilize a measuring window width of 4 for segment comparison when evaluating split candidates. Furthermore, distribution-based segmentation relies on a predefined optimal number of segments. Following the study's suggestion, we use the formula $\min(\max(10, n / 2), 30)$, where $n$ represents the case length. For both real-world event logs, this results in 15 segments for a 30-event trace prefix. We utilize the same KernelSHAP computation across all differently defined segments. All methods provide explanations for the same feature space, encompassing both control-flow and time-related features. Finally, we note that per-event attribution operates at a finer granularity than the segment-level methods.

\paragraph{Evaluation of Segmentation Quality.} 
% Our segmentation algorithm partitions traces into coherent subsequences based on transition probabilities. 
We evaluate segmentation quality differently for synthetic and real-world event logs. For the synthetic log, ground-truth change points are available. Therefore, as evaluation metric we employ the RandIndex, Precision, Recall, and F-1 score \cite{truong2020selective}. These metrics assess how accurately our segmentation algorithm identifies the true segment boundaries.
For real-world logs, no ground-truth is available. Therefore, we report descriptive statistics on the resulting segmentation, namely the number and lengths of segments to reflect granularity. Furthermore, we compute the proportion of traces covered by the top-10 most frequent change point patterns as follows. For each trace with change points $\tau_1, \ldots, \tau_k$, we extract the transition boundary $(a_{\tau_i}, a_{\tau_i+1})$ at each detected change point. After computing the frequency of unique patterns, we report the top-10 patterns coverage \textemdash how consistent the segmentation is across traces. 
We also evaluate segmentation quality using entropic relevance (ER)~\cite{polyvyanyy2020entropic}, an information-theoretic metric~\cite{shannon1948mathematical}. ER quantifies the number of bits needed to encode traces based on the directly-follows transition probabilities within each segment. Segments with more deterministic transitions yield lower ER scores, indicating that the segmentation captures coherent patterns that require fewer bits to describe. The implementation was adopted from the work~\cite{peeperkorn2026model}.

\paragraph{Evaluation of Feature Attributions.} 
In the absence of ground-truth explanations, feature attribution methods are commonly evaluated through perturbation-based metrics \cite{agarwal2022openxai,baer2025class,turbe2023evaluation}. Following this idea, we assess the faithfulness of segment-level SHAP values using Relative Prediction Change (RPC). We compute RPC for both important (RPCI) and unimportant (RPCU) segments, where importance is defined by the absolute SHAP value. The top-$k$ segments are selected such that they cover 40\% of the trace prefix $\sigma$, which provides a large enough perturbation to affect the prediction without fully destroying the trace structure. Intuitively, RPC measures how much the model prediction changes after perturbing selected segments. A faithful explanation should assign high importance to segments whose perturbation strongly affects the prediction, and low importance to segments whose perturbation has little effect. To compute RPCI, we perturb the top-$k$ most important segments by replacing their feature values with mean values from 100 randomly sampled training cases. RPCU is computed by perturbing the segments deemed unimportant while keeping the top-$k$ fixed. We report both metrics per predicted class to account for class-dependent effects \cite{baer2025class}.

\section{Experimental Evaluation}
\label{sec:results}

This study investigates how meaningful segment-level feature attributions can be derived for predictive process monitoring. We address this by exploring the following research questions:(\textbf{RQ1}) how faithful are segment-level SHAP value explanations to the model’s output compared to non-segment-level SHAP values? (\textbf{RQ2}) Can SHAP values identify meaningful segments within a case that are most influential for a given prediction?

\subsection{Trace Segmentation}

The results of the segmentation quality performed on the synthetic data are shown in Table \ref{tab:segm_quality_synth}. The agreement between predicted and ground truth change points, as quantified by the RandIndex, exceeds 0.9 for both canceled and accepted cases with one and two change points, correspondingly. Segmentation accuracy is slightly higher for the former due to the smaller number of segments, achieving an F1 score of 89\%.  

% \vspace{-10pt}

\begin{table}[!ht]
\centering
\caption{Segmentation quality for the synthetic dataset with the activities specified as change points.}
\label{tab:segm_quality_synth}
\scriptsize
\resizebox{\columnwidth}{!}{%
\begin{tabular}{l| c c c c}
% \toprule
\toprule
 & \multirow{2}{*}{\textbf{RandIndex}} & \multirow{2}{*}{\textbf{Precision}} & 
 \multirow{2}{*}{\textbf{Recall}} & 
 \multirow{2}{*}{\textbf{F1 score}} \\
\textbf{Sample (change points)} &  &  &  &  \\ 
\midrule
canceled cases (``skipped\_hq") & 0.928 & 0.875 & 0.925 & 0.885 \\
accepted cases (``contact\_hq", ``calculate\_offer”) & 0.951 & 0.8582 & 0.906 & 0.874 \\ 
\bottomrule
\end{tabular}
}

\end{table}
 
Regarding the real logs, table \ref{tab:seg_quality_real} summarizes results for BPIC17 and BPIC15. In the BPIC17, correctly predicted cases contain, on average, three events per segment and approximately eight segments per prefix. The ten most frequent segmentation patterns for True Negative (TN) and False Negative (FN) samples represent the control flow of 81\% and 77\% of all trace prefixes in the test set.
In contrast, canceled cases follow more diverse control-flow paths; the 212 identified patterns result in the ten most frequent patterns covering only 42\% of the traces. Finally, the application of the control-flow-aware segmentation algorithm reduces the entropic relevance from 37.81 to 6.68, indicating that the identified segments yield a substantially more compressible, and thus more structured, trace representation.       

The BPIC15 event log contains 283 unique activities characterized by low or near-zero transition probabilities between the majority of event pairs. This high log variability, which is also supported by the high ER of 100, causes a more granular segmentation of the trace prefixes, resulting in approximately 13 segments with an average of two events each. These results are distributed uniformly across all samples, which does not allow the identification of any distinct process behaviors. Consequently, the results indicate that each prefix exhibits a unique, case-specific segmentation pattern.         

\vspace{-10pt}

\begin{table}[!ht]
\centering
\caption{Segmentation evaluation on the real-world datasets.}
\label{tab:seg_quality_real}
\resizebox{\columnwidth}{!}{%
\begin{tabular}{l|c|c|c|c|c|c|c|c}
\hline
                      & \multicolumn{4}{c|}{$\mathbf{BPIC17}$} & \multicolumn{4}{c}{$\mathbf{BPIC15}$} \\ \cline{2-9} 
                      & TP       & FP     & TN       & FN      & TP      & FP      & TN       & FN     \\ \hline
Sample size           & 705      & 4      & 2004     & 288     & 45      & 7       & 125      & 1      \\
Number of segments &
  $8.93 \pm 1.26$&
  $6.25 \pm 2.05$ &
  $8.21 \pm 1.16$ &
  $7.72 \pm 1.56$ &
  $13.31 \pm 0.69$ &
  $12.86 \pm 0.64$ &
  $13.50  \pm 0.74$ &
  $14 \pm 0.0$ \\
Segment length &
  $3.44 \pm 0.61$ &
  $5.62 \pm 2.58$ &
  $3.74 \pm 0.70$ &
  $4.14 \pm 1.33$ &
  $2.26 \pm 0.125$ &
  $2.34 \pm 0.12$ &
  $2.23 \pm 0.13$ &
  $2.14 \pm 0.0$ \\
\begin{tabular}[c]{@{}l@{}}Top-10 patterns\\ coverage\end{tabular} &
  $42.1\%$ &
  $100\%$ &
  $81.4\%$ &
  $77.4\%$ &
  $22.2\%$ &
  $100\%$ &
  $8\%$ &
  $100\%$ \\
Number of patterns    & 212      & 3      & 49       & 115     & 45      & 7       & 125      & 1      \\ \hline
ER full log           & \multicolumn{4}{c|}{37.81 $\pm$ 6.11}  & \multicolumn{4}{c}{100.1 $\pm$ 10.36} \\
ER after segmentation & \multicolumn{4}{c|}{ 6.68 $\pm$ 2.50}   & \multicolumn{4}{c}{10.93 $\pm$ 2.49}  \\ \hline
\end{tabular}%
}
\end{table}

\subsection{Evaluation of Feature Attributions}

For the quantitative evaluation of segment-level SHAP explanations, we use perturbation-based metrics described in Section \ref{sec:setup}. Table \ref{tab:sv_quality} reports RPCI and RPCU for the three segmentation strategies on BPIC17.  
A faithful explanation yields high RPCI (important segments substantially affect the prediction) and low RPCU (unimportant segments have minimal effect). 
The parameter $k$ differs across strategies since each produces segments of varying lengths. To ensure a fair comparison, $k$ is chosen such that the perturbed segments cover 40\% of the trace length.  
The following interpretation of the results addresses our \textbf{RQ1}. All three segmentation strategies yield faithful explanations, with higher RPCI and lower RPCU for class 1. RPCI exceeds 1.0 for class 0 cases because their near-zero scores amplify changes when important segments are perturbed, still indicating high faithfulness (e.g., removing completion/validation segments shifts predictions toward cancellation). For class 1, where scores are closer to 1.0, RPCI stays below 1.0 with lower variance, reflecting more consistent explanations across strategies. 

The distribution-based and control-flow-aware segmentation strategies sacrifice some faithfulness, which can be attributed to the aggregation of events into semantically coherent segments. In contrast, event-level attributions can precisely target the most individually contributing events. However, it should be noted that RPCU for class 0 decreases as segmentation granularity becomes coarser in the both event logs. For event-level attribution, perturbing individual low-ranked events might disrupt local dependencies between neighboring events, resulting unexpected prediction changes. Segment-level strategies avoid this by perturbing coherent groups of events together, preserving internal structure and producing a cleaner separation between important and unimportant parts of the trace.    

\vspace{-10pt}
\setlength{\tabcolsep}{5pt}

\begin{table}[!ht]
\centering
\caption{Evaluation of faithfulness of SHAP value explanations.}
\label{tab:sv_quality}
\resizebox{\columnwidth}{!}{%
\begin{tabular}{l| c c c| c c c}

\toprule
& \multicolumn{3}{c|}{BPIC17} & \multicolumn{3}{c}{BPIC15} \\
\midrule
\textbf{Sample} & $k$ & $\mathrm{RPCI}$ & $\mathrm{RPCU}$ & $k$ & $\mathrm{RPCI}$ & $\mathrm{RPCU}$  \\ 
\midrule
\multicolumn{7}{l}{\textit{Per-event segmentation}} \\ 
\midrule
Pred. class 0 & \multirow{2}{*}{12} & 0.909 $\pm$ 0.856 & 0.525 $\pm$ 0.382 & \multirow{2}{*}{12} & 2.140 $\pm$ 2.191 & 0.311 $\pm$ 0.179 \\
Pred. class 1 &  & \textbf{0.832} $\pm$ \textbf{0.097} & \textbf{0.005} $\pm$ \textbf{0.012} & & \textbf{0.973 }$\pm$\textbf{ 0.016} & \textbf{0.004} $\pm$ \textbf{0.005} \\

\midrule
\multicolumn{7}{l}{\textit{Distribution-based segmentation}} \\ \midrule
Pred. class  0 & \multirow{2}{*}{5} & 1.352 $\pm$ 0.860 & 0.329 $\pm$ 0.290 & \multirow{2}{*}{5} & 2.139 $\pm$ 2.235 & 0.304 $\pm$ 0.166 \\
Pred. class  1 &  &  0.773 $\pm$ 0.157 & 0.005 $\pm$ 0.013 & & 0.951 $\pm$ 0.049 & 0.018 $\pm$ 0.037 \\

\midrule
\multicolumn{7}{l}{\textit{Control-flow-aware segmentation}} \\ 
\midrule
Pred. class 0 &  \multirow{2}{*}{5} & 1.278 $\pm$ 0.777 & \textbf{0.233} $\pm$ \textbf{0.230} & \multirow{2}{*}{6} & 2.154 $\pm$ 2.126 & \textbf{0.242} $\pm$ \textbf{0.148}\\
Pred. class 1 & & 0.769 $\pm$ 0.113 & 0.008 $\pm$ 0.018 &  & 0.972 $\pm$ 0.017  & 0.007 $\pm$ 0.015\\

\bottomrule
\end{tabular}%
} 
\end{table}

\subsection{Segment-level Explanations}
 
\begin{figure}[!ht]
    \centering
    \makebox[\textwidth][c]{%
    \includegraphics[width=1.1\textwidth]{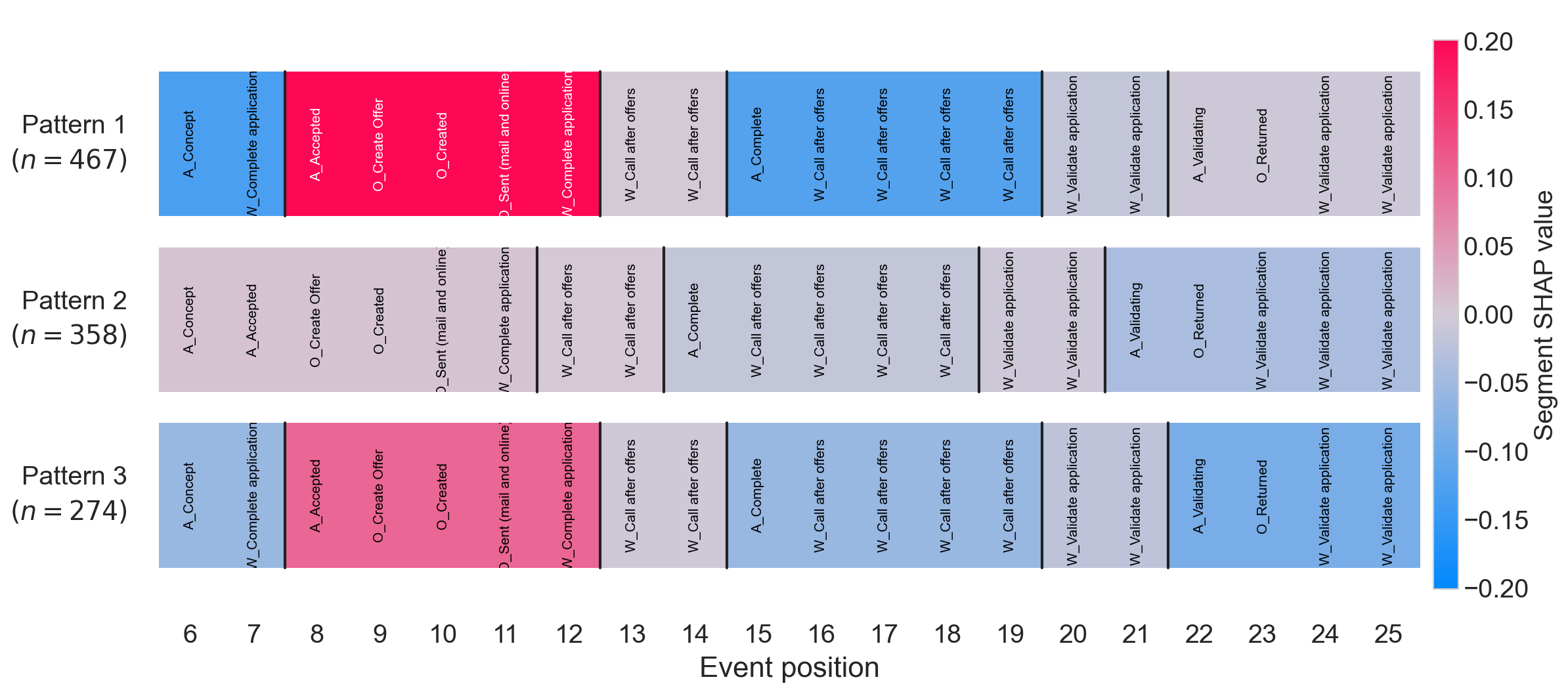}
   }
   \caption{Visualization of SHAP values for representative traces from the three most frequent segmentation patterns regarding \textit{accepted} cases for the BPIC17.}
    \label{fig:seg_sv_tn}
\end{figure}

\begin{figure}[!ht]
    \centering
    \makebox[\textwidth][c]{%
    \includegraphics[width=1.1\textwidth]{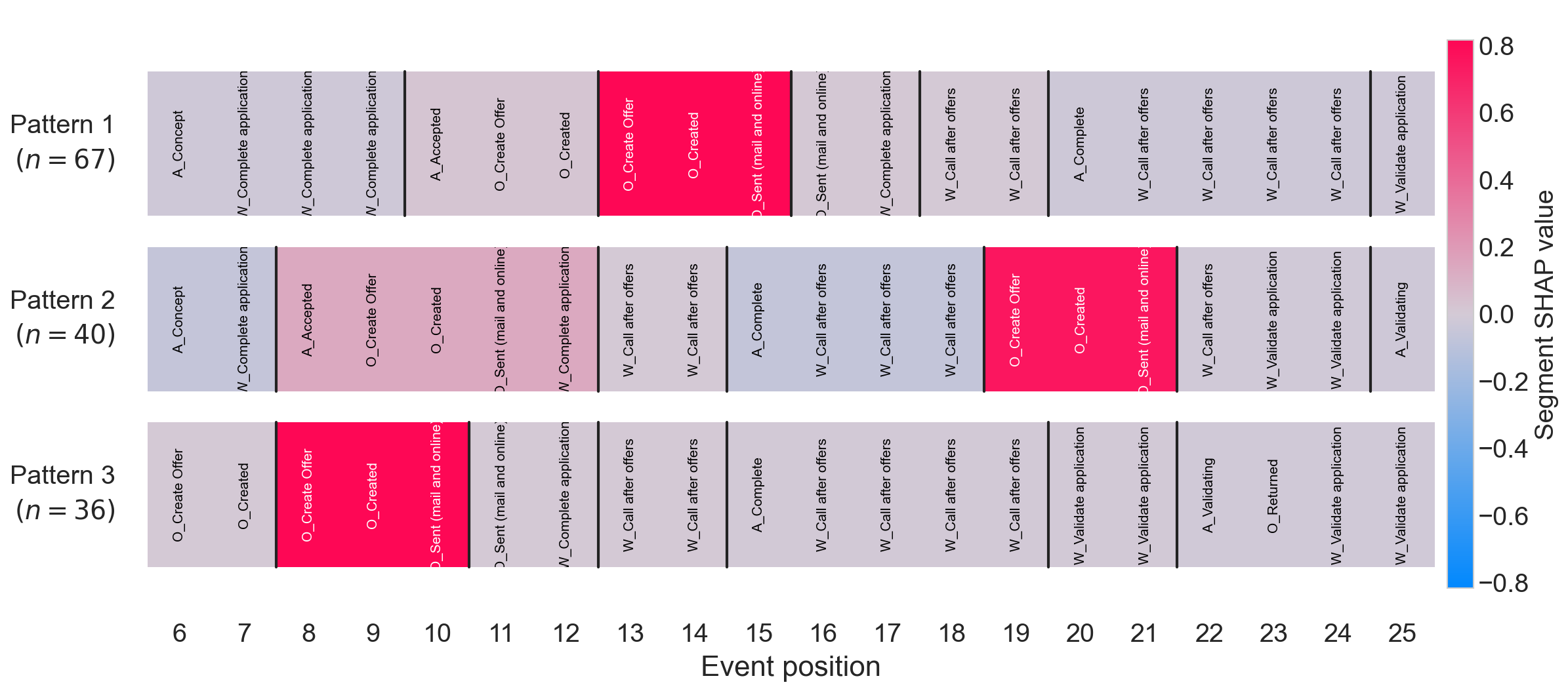}
   }
   \caption{Visualization of SHAP values for representative traces from the three most frequent segmentation patterns regarding \textit{cancelled} cases for the BPIC17.}
    \label{fig:seg_sv_tp}
\end{figure}

\begin{figure}[!ht]
    \centering
    % --- First Plot ---
    \begin{subfigure}{\linewidth}
        \centering
        \makebox[\textwidth][c]{%
        \includegraphics[width=1.1\textwidth]{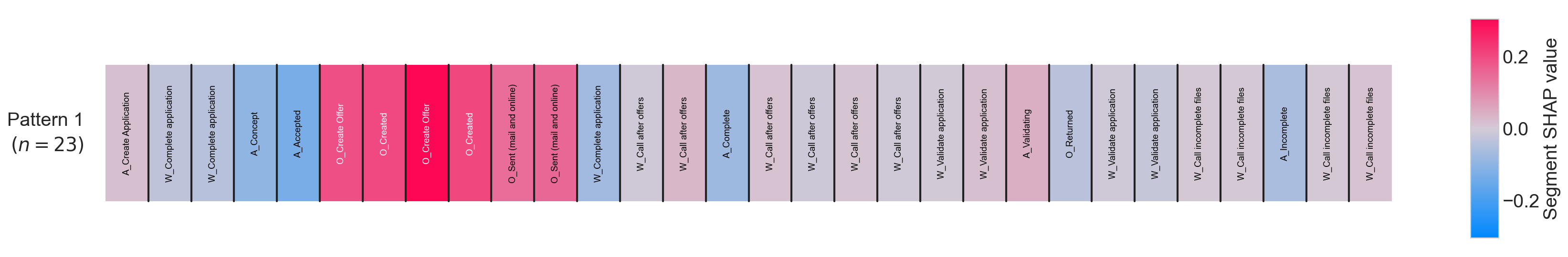}}
        \label{fig:tp_per_event}
    \end{subfigure}
    \begin{subfigure}{\linewidth}
        \centering
        \makebox[\textwidth][c]{%
        \includegraphics[width=1.1\textwidth]{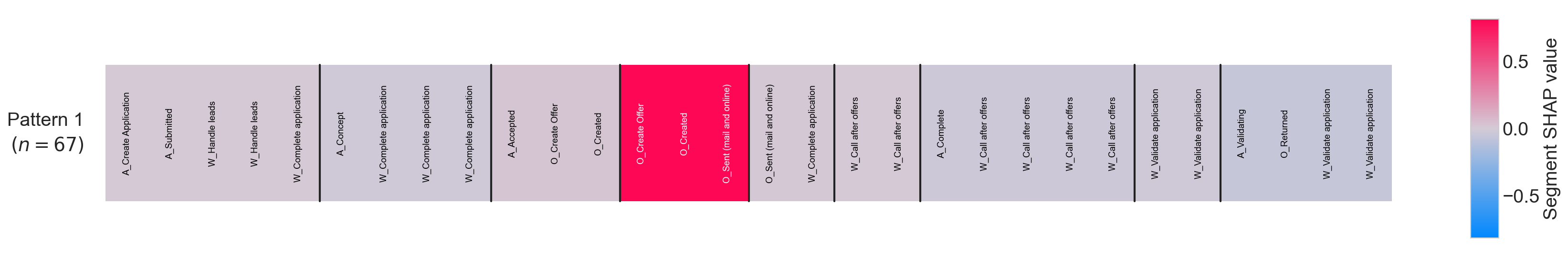}}
        \label{fig:tp_transition}
    \end{subfigure}

    \caption{Comparison of event-level and segment-level SHAP value explanations for the BPIC17 log.}
    \label{fig:seg_comparison_bpi17}
\end{figure}

In this analysis to address our \textbf{RQ2}, we consider the three most frequent segmentation patterns we generated from BPIC17, and the assigned SHAP values for each pattern and each class.
In~\autoref{fig:seg_sv_tn}, we illustrate the accepted cases. As previously discussed in the segmentation analysis, these cases follow more regular and recurring control-flow patterns, which indicates a more stable process execution. In these cases, segments containing transitions such as \texttt{A\_Concept} $\rightarrow$ \texttt{W\_Complete\_application} consistently push the prediction toward acceptance, as reflected by negative SHAP values. The same holds for later segments related to application completion and validation, such as \texttt{A\_Complete}. This suggests that the model associates smooth and structured progression through the process with a favorable outcome.
On the other hand, in~\autoref{fig:seg_sv_tp}, the canceled cases show a different pattern, mainly driven by repeated offer creation. In particular, segments related to multiple created offers receive high positive SHAP values and increasingly push the prediction toward cancellation. The last created offer often has the strongest effect. For stakeholders, this is meaningful because it points to a potential bottleneck detection: repeated offer generation is strongly associated with unsuccessful case outcomes. A visual comparison of event-level and segment-level feature attributions for these cases is provided in~\autoref{fig:seg_comparison_bpi17}.

We also analyze the wrongly predicted cases, which further support this interpretation: accepted cases wrongly predicted as canceled often already contain multiple created offers in the observed prefix, whereas canceled cases wrongly predicted as accepted lacked any control-flow activities indicating an application completion or validation stage. This is likely influenced by the 30-event prefix, since decisive outcome-related behavior may appear later in the trace. 

Overall, these findings show the value of segment-level explanations: instead of attributing predictions to isolated events or to the whole trace, they reveal which \emph{part of the process} drives the prediction and contextualizes particular activities in their segments.

\vspace{-10pt}

\vspace{-20pt}

%\paragraph{Analysis of the BPIC15.} The BPIC15 event log is characterized by high variability due to the large number of activities with low transition probabilities. This is reflected in the segmentation analysis, where each trace exhibits a unique segmentation pattern. Such variability complicates the general conclusions that can be drawn from segment-level explanations about model behavior.The SHAP values of cases satisfying the defined LTL rule (TP sample) indicate higher importance for the segment containing the activity ``retrieve missing data". However, segments containing activities functionally related to missing data retrieval (e.g., ``send email retrieve missing data") also receive comparably high attribution magnitudes. Since activities were encoded as integer identifiers with no access to their textual labels, this pattern suggests that the model has learned similar internal representations for activities that serve functionally equivalent stages in the process. From a prediction standpoint, these activities might signal the same underlying process state, namely the presence of unresolved missing information, which is strongly associated with the case outcome. However, for the cases that violate the LTL rule (TN sample), we observe a common pattern in model behavior. The SHAP values of the last two or three segments of each prefix receive the strongest negative attributions, which signifies that the model has not encountered the required activity or functionally equivalent activities in the current control flow. 

\section{Discussion and Limitations}
\label{sec:discussion}

This study proposes a method to provide explainability of sequential models in the PPM tasks through  the estimation of SHAP values at the segment level. The control-flow-aware segmentation algorithm allows trace partitioning into segments preserving structural dependencies. Our experiments show that event logs with moderate control-flow variability, such as BPIC17, exhibit clearly identifiable phases: transition shifts are well-defined due to the log structure and a manageable number of unique activities, allowing the segmentation algorithm to produce meaningful partitions. Aggregating the resulting segmentation patterns across traces already facilitates a high-level analysis of the event log, revealing common phase structures without requiring inspection of individual traces. 

Through the quantitative evaluation of the obtained segment-level feature attributions, we demonstrated that regions meaningful to the model can be identified. Across both datasets and all segmentation strategies, perturbing the top-attributed segments causes substantial prediction changes, while perturbing low-attributed segments has minimal effect. This confirms that the identified regions are influential rather than arbitrary.

Comparing our approach against two alternative segmentation strategies, we demonstrated that grouping activities based on control-flow information leads to more comprehensible explanations that allow the identification of distinct transition patterns. While event-level feature attributions are more faithful to the model's behavior due to their finer granularity, segment-level attributions enable the analysis of model behavior at a coarser level. In particular, they mitigate the interleaving of important and unimportant events, thereby increasing the overall comprehensibility of the explanations.

Since our segmentation method relies on the directly-follows relations among activities, low transition probabilities lead to fine-grained segmentation, which limits the aggregation of segmentation patterns across cases. High variability in the event log control flow further exacerbates this issue, as few cases share similar segment structures, making it difficult to draw generalizable conclusions about model behavior at the segment level. Conversely, event logs with fewer distinct activities and short trace prefixes offer limited benefit from segmentation, as there is insufficient sequential structure to produce meaningful groupings.
\section{Conclusion}
\label{sec:conclusion}

In this paper, we proposed a feature attribution-based explainability method for PPM that preserves the control-flow perspective of business processes. More specifically, we introduced a control-flow-aware segmentation algorithm that partitions trace prefixes into segments based on shifts in directly-follows transition probabilities, and we used these segments as explanatory units for the computations of segment-level SHAP values. The results on synthetic data showed that the proposed segmentation method is capable of recovering meaningful change points with high agreement to ground truth, while the experiments on real-world event logs showed that it produces interpretable segmentation and faithful explanations. In particular, for BPIC17, the method revealed recurring trace segments associated with accepted and canceled cases, making it possible to relate model predictions to concrete parts of the process rather than to isolated events of fully aggregated traces.

Overall, our findings show that segment-level explanations offer a useful middle ground between fine-grained event-level and trace-level explanations. Although the former remains slightly more faithful due to its finer granularity and simplicity, segment-level explanations improve comprehensibility by preserving local control-flow context and reducing the interleaving of important and unimportant events. At the same time, the study highlights that the benefit of the proposed method depends on the structure of the event log: logs with moderate control-flow variability allow the discovery of meaningful and comparable segments, whereas highly variable logs lead to very fine-grained and case-specific segmentation. In future work, we will extend the method beyond binary outcome prediction enabling analysis of distinct failure modes and supporting more targeted decision-making. 

{\bfseries Disclosure of Interests.} The authors have no competing interests to declare that are relevant to the content of this article.

\bibliographystyle{splncs04}
\bibliography{bibliography}

@inproceedings{tax2017predictive,
  title={Predictive business process monitoring with LSTM neural networks},
  author={Tax, Niek and Verenich, Ilya and La Rosa, Marcello and Dumas, Marlon},
  booktitle={CAiSE},
    year={2017},
    doi={10.1007/978-3-319-59536-8_30},
}

@inproceedings{weytjens2020process,
  title={Process outcome prediction: CNN vs. LSTM (with attention)},
  author={Weytjens, Hans and De Weerdt, Jochen},
  booktitle={BPM},
    year={2020},
    doi={10.1007/978-3-030-66498-5_24},
}

@article{stevens2025generating,
  title={Generating feasible and plausible counterfactual explanations for outcome prediction of business processes},
  author={Stevens, Alexander and Ouyang, Chun and De Smedt, Johannes and Moreira, Catarina},
  journal={IEEE TSC},
  year={2025},
  doi={10.1109/TSC.2025.3609837}
  }

@article{buliga2025guiding,
  title={Guiding the generation of counterfactual explanations through temporal background knowledge for predictive process monitoring},
  author={Buliga, Andrei and Di Francescomarino, Chiara and Ghidini, Chiara and Donadello, Ivan and Maggi, Fabrizio Maria},
  journal={Data Mining and Knowledge Discovery},
    year={2025},
    doi={10.1007/s10618-025-01117-3},
}

@inproceedings{rizzi2020explainability,
  title={Explainability in predictive process monitoring: When understanding helps improving},
  author={Rizzi, Williams and Di Francescomarino, Chiara and Maggi, Fabrizio Maria},
  booktitle={BPM},
    year={2020},
    doi={10.1007/978-3-030-58638-6_9},
}

@Article{elkhawaga_2022_xai,
AUTHOR = {El-khawaga, Ghada and Abu-Elkheir, Mervat and Reichert, Manfred},
TITLE = {XAI in the Context of Predictive Process Monitoring: An Empirical Analysis Framework},
JOURNAL = {Algorithms},
YEAR = {2022},
}

@inproceedings{velmurugan2021evaluating,
  title={Evaluating stability of post-hoc explanations for business process predictions},
  author={Velmurugan, Mythreyi and Ouyang, Chun and Moreira, Catarina and Sindhgatta, Renuka},
  booktitle={ICSOC},
    year={2021},
}

@inproceedings{galanti2020explainable,
  title={Explainable predictive process monitoring},
  author={Galanti, Riccardo and Coma-Puig, Bernat and de Leoni, Massimiliano and Carmona, Josep and Navarin, Nicol{\`o}},
  booktitle={ICPM},
    year={2020},
}

@inproceedings{sundararajan2020many,
  title={The many Shapley values for model explanation},
  author={Sundararajan, Mukund and Najmi, Amir},
  booktitle={International conference on machine learning},
    year={2020},
    doi={10.48550/arXiv.1911.06292},
}

@inproceedings{jiang2024seqshap,
  title={SeqSHAP: subsequence level shapley value explanations for sequential predictions},
  author={Jiang, Guanyu and Zhuang, Fuzhen and Song, Bowen and Zhu, Yongchun and Sun, Ying and Wang, Weiqiang and Wang, Deqing},
  booktitle={DASFAA},
    year={2024},
}

@inproceedings{lundberg_unified_2017,
author = {Lundberg, Scott M. and Lee, Su-In},
title = {A unified approach to interpreting model predictions},
year = {2017},
isbn = {9781510860964},
publisher = {Curran Associates Inc.},
address = {Red Hook, NY, USA},
booktitle = {Proceedings of the 31st International Conference on Neural Information Processing Systems},
pages = {4768–4777},
numpages = {10},
location = {Long Beach, California, USA},
series = {NIPS'17},
doi={10.48550/arXiv.1705.07874},
}

@inproceedings{bento_timeshap_2021,
author = {Bento, Jo\~{a}o and Saleiro, Pedro and Cruz, Andr\'{e} F. and Figueiredo, M\'{a}rio A.T. and Bizarro, Pedro},
title = {TimeSHAP: Explaining Recurrent Models through Sequence Perturbations},
year = {2021},
publisher = {Association for Computing Machinery},
address = {New York, NY, USA},
doi = {10.1145/3447548.3467166},
booktitle = {Proceedings of the 27th ACM SIGKDD Conference on Knowledge Discovery \& Data Mining},
pages = {2565–2573},
numpages = {9},
location = {Virtual Event, Singapore},
series = {KDD '21}
}

@inproceedings{agarwal2022openxai,
author = {Agarwal, Chirag and Krishna, Satyapriya and Saxena, Eshika and Pawelczyk, Martin and Johnson, Nari and Puri, Isha and Zitnik, Marinka and Lakkaraju, Himabindu},
title = {OpenXAI: towards a transparent evaluation of post hoc model explanations},
year = {2022},
isbn = {9781713871088},
publisher = {Curran Associates Inc.},
address = {Red Hook, NY, USA},
booktitle = {Proceedings of the 36th International Conference on Neural Information Processing Systems},
articleno = {1148},
numpages = {16},
location = {New Orleans, LA, USA},
series = {NIPS '22}
}

@article{turbe2023evaluation,
  title={Evaluation of post-hoc interpretability methods in time-series classification},
  author={Turb{\'e}, Hugues and Bjelogrlic, Mina and Lovis, Christian and Mengaldo, Gianmarco},
  journal={Nature Machine Intelligence},
  volume={5},
  number={3},
    year={2023},
    doi={10.1038/s42256-023-00620-w},
}

@inproceedings{demoor2025simbank,
  title={Simbank: from simulation to solution in prescriptive process monitoring},
  author={De Moor, Jakob and Weytjens, Hans and De Smedt, Johannes and De Weerdt, Jochen},
  booktitle={BPM},
    year={2025},
    doi={10.1007/978-3-032-02929-4_10},
}

@article{nayebi2023window,
title = {WindowSHAP: An efficient framework for explaining time-series classifiers based on Shapley values},
journal = {Journal of Biomedical Informatics},
volume = {144},
pages = {104438},
year = {2023},
issn = {1532-0464},
author = {Amin Nayebi and Sindhu Tipirneni and Chandan K. Reddy and Brandon Foreman and Vignesh Subbian},
}

@article{killick2012optimal,
  title={Optimal detection of changepoints with a linear computational cost},
  author={Killick, Rebecca and Fearnhead, Paul and Eckley, Idris A},
  journal={Journal of the American Statistical Association},
  volume={107},
  number={500},
    year={2012},
    doi={10.1080/01621459.2012.737745},
}

@inproceedings{sivill2022limesegment,
  title={Limesegment: Meaningful, realistic time series explanations},
  author={Sivill, Torty and Flach, Peter},
  booktitle={International Conference on Artificial Intelligence and Statistics},
    year={2022},
}

@inproceedings{schlegel2021ts,
  title={Ts-mule: Local interpretable model-agnostic explanations for time series forecast models},
  author={Schlegel, Udo and Vo, Duy Lam and Keim, Daniel A and Seebacher, Daniel},
  booktitle={ECML-PKDD},
    year={2021},
    doi={10.1007/978-3-030-93736-2_40},
}

@inproceedings{baer2025class,
  title={Class-dependent perturbation effects in evaluating time series attributions},
  author={Baer, Gregor and Grau, Isel and Zhang, Chao and Van Gorp, Pieter},
  booktitle={XAI World Conference},
    year={2025},
}

@inproceedings{polyvyanyy2020entropic,
  title={An entropic relevance measure for stochastic conformance checking in process mining},
  author={Polyvyanyy, Artem and Moffat, Alistair and Garc{\'\i}a-Ba{\~n}uelos, Luciano},
  booktitle={ICPM},
    year={2020},
    doi={10.1109/ICPM49681.2020.00024},
}

@article{shannon1948mathematical,
  title={A mathematical theory of communication},
  author={Shannon, Claude Elwood},
  journal={The Bell system technical journal},
  volume={27},
  number={3},
    year={1948},
}

@article{teinemaa2019outcome,
  title={Outcome-oriented predictive process monitoring: Review and benchmark},
  author={Teinemaa, Irene and Dumas, Marlon and Rosa, Marcello La and Maggi, Fabrizio Maria},
  journal={ACM TKDD},
  volume={13},
  number={2},
    year={2019},
    doi={10.1145/3301300},
}

@article{truong2020selective,
  title={Selective review of offline change point detection methods},
  author={Truong, Charles and Oudre, Laurent and Vayatis, Nicolas},
  journal={Signal processing},
  volume={167},
    year={2020},
    doi={10.1016/j.sigpro.2019.107299},
}

@article{li2024automatic,
  title={Automatic change-point detection in time series via deep learning},
  author={Li, Jie and Fearnhead, Paul and Fryzlewicz, Piotr and Wang, Tengyao},
  journal={Journal of the Royal Statistical Society Series B: Statistical Methodology},
  volume={86},
  number={2},
    year={2024},
}

@article{deldari2020espresso,
  title={Espresso: Entropy and shape aware time-series segmentation for processing heterogeneous sensor data},
  author={Deldari, Shohreh and Smith, Daniel V and Sadri, Amin and Salim, Flora},
  journal={Proceedings of the ACM on Interactive, Mobile, Wearable and Ubiquitous Technologies},
  volume={4},
  number={3},
    year={2020},
    doi={10.1145/3411832},
}

@article{hamed2026changepoint,
  title={Changepoint detection as a light data-driven approach to battery state-of-health prediction},
  author={Hamed, Hamid and Reis, Albin Conde and Choobar, Behnam Ghalami and Pang, Quanquan and Killick, Rebecca and Safari, Mohammadhossein},
  journal={Cell Reports Physical Science},
  year={2026},
    doi={10.1016/j.xcrp.2026.103157},
}

@article{aas2021explaining,
  title={Explaining individual predictions when features are dependent: More accurate approximations to Shapley values},
  author={Aas, Kjersti and Jullum, Martin and L{\o}land, Anders},
  journal={Artificial Intelligence},
  volume={298},
    year={2021},
    doi={10.1016/j.artint.2021.103502},
}

@inproceedings{crabbe2021explaining,
  title={Explaining time series predictions with dynamic masks},
  author={Crabb{\'e}, Jonathan and Van Der Schaar, Mihaela},
  booktitle={ICML},
  year={2021},
    doi={10.48550/arXiv.2106.05303},
}

@inproceedings{
liu2024explaining,
title={Explaining Time Series via Contrastive and Locally Sparse Perturbations},
author={Zichuan Liu and Yingying Zhang and Tianchun Wang and Zefan Wang and Dongsheng Luo and Mengnan Du and Min Wu and Yi Wang and Chunlin Chen and Lunting Fan and Qingsong Wen},
booktitle={ICLR},
year={2024},
}

@inproceedings{doddaiah2022class,
  title={Class-specific explainability for deep time series classifiers},
  author={Doddaiah, Ramesh and Parvatharaju, Prathyush and Rundensteiner, Elke and Hartvigsen, Thomas},
  booktitle={ICDM},
  year={2022},
}

@article{peeperkorn2026model,
  title={Model-driven stochastic trace clustering},
  author={Peeperkorn, Jari and De Smedt, Johannes and De Weerdt, Jochen},
  journal={Information Systems},
  volume={139},
  year={2026},
  publisher={Elsevier},
  doi={10.1016/j.is.2026.102697},
}

\end{document}